\documentclass[10pt,twocolumn,letterpaper]{article}

\usepackage{cvpr}              

\usepackage{times}
\usepackage{epsfig}
\usepackage{graphicx}
\usepackage{amsmath}
\usepackage{amssymb}
\usepackage{multicol}
\usepackage{multirow}
\usepackage{latexsym}
\usepackage{pifont}
\usepackage{fdsymbol}
\usepackage{marvosym}
\usepackage{flushend}
\usepackage{mathtools}
\usepackage[accsupp]{axessibility} 

\usepackage[dvipsnames]{xcolor}

\newcommand{\green}[1]{{\color{green}#1}}
\newcommand{\blue}[1]{{\color{blue}#1}}
\newcommand{\orange}[1]{{\color{orange}#1}}

\usepackage{threeparttable}
\usepackage{etoolbox}
\AtBeginEnvironment{tablenotes}{\footnotesize}
\usepackage{multirow}
\usepackage{amssymb}
\usepackage{pifont}
\newcommand{\cmark}{\ding{51}}%
\newcommand{\xmark}{\ding{55}}%
\usepackage{makecell}
\usepackage{algorithm}
\usepackage{algpseudocode}
\usepackage{amsmath}
\usepackage{booktabs}

\definecolor{cvprblue}{rgb}{0.21,0.49,0.74}
\usepackage[pagebackref,breaklinks,colorlinks,citecolor=cvprblue]{hyperref}

\def\paperID{9066} 
\def\confName{CVPR}
\def\confYear{2024}

\title{DIBS: Enhancing Dense Video Captioning with Unlabeled Videos via \\  Pseudo Boundary Enrichment and Online Refinement}

\author{Hao Wu\textsuperscript{1,2 $\clubsuit$},\quad Huabin Liu\textsuperscript{2,3 $\clubsuit$},\quad Yu Qiao\textsuperscript{2},\quad Xiao Sun\textsuperscript{2 \Letter} \\
\textsuperscript{1} University of Science and Technology of China~\quad
\textsuperscript{2} Shanghai Artificial Intelligence Laboratory\\
\textsuperscript{3} Shanghai Jiao Tong University  \\
{\tt\small wuhao@mail.ustc.edu.cn,~huabinliu@sjtu.edu.cn,~\{qiaoyu, sunxiao\}@pjlab.org.cn}
}

\begin{document}
\maketitle
\let\thefootnote\relax\footnotetext{$\clubsuit$\ Co-first authors. Work done as interns at Shanghai AI Laboratory.}
\let\thefootnote\relax\footnotetext{\Letter\ Corresponding author.}

\begin{abstract}
We present \textbf{D}ive \textbf{I}nto the \textbf{B}oundarie\textbf{S} (\textbf{DIBS}), a novel pretraining framework for dense video captioning (DVC), that elaborates on improving the quality of the generated event captions and their associated pseudo event boundaries from unlabeled videos. By leveraging the capabilities of diverse large language models (LLMs), we generate rich DVC-oriented caption candidates and optimize the corresponding pseudo boundaries under several meticulously designed objectives, considering diversity, event-centricity, temporal ordering, and coherence. Moreover, we further introduce a novel online boundary refinement strategy that iteratively improves the quality of pseudo boundaries during training. Comprehensive experiments have been conducted to examine the effectiveness of the proposed technique components. By leveraging a substantial amount of unlabeled video data, such as HowTo100M~\cite{miech19howto100m}, we achieve a remarkable advancement on standard DVC datasets like YouCook2~\cite{ZhXuCo18youcook2} and ActivityNet~\cite{krishna2017anet}. We outperform the previous state-of-the-art Vid2Seq~\cite{Vid2Seq2023} across a majority of metrics, achieving this with just 0.4\% of the unlabeled video data used for pre-training by Vid2Seq.


\end{abstract}
\section{Introduction}
\label{sec:intro}

Dense video captioning (DVC), a challenging task in video understanding, involves the temporal localization and captioning of all events within an untrimmed video~\cite{DVC2017}. Compared to standard video captioning that generates a single caption for a short video clip~\cite{chen2018less,zhang2020object,xu2018dual}, the complexity of dense captioning significantly increases as it requires localizing multiple events in long-term video sequences and much more detailed captioning. 

Particularly, the event boundary is paramount in dense video captioning. It provides precise event localization, ensuring the generated captions are accurate, coherent, and contextually relevant. Unfortunately, data containing precise event boundaries is rare and expensive to annotate. This scarcity poses a substantial performance bottleneck.

Pioneering efforts have been made to address data shortage challenges in DVC. Notably, several weakly supervised approaches~\cite{WSDEC2018,EC-SL2021} have endeavored to approximate fully supervised performance without relying on all the  boundary annotations provided by existing datasets, thereby circumventing the need for such annotations. 
However, these methods take a conservative approach to the problem by carefully re-designing the DVC training or testing framework, aiming to theoretically develop self-sufficient techniques and reduce the reliance on precise boundary annotations. They have not yet substantially incorporated larger-scale data for training purposes to drive improved performance. Instead, we take a direct approach to address the fundamental data scarcity issues. Specifically, we introduce an effective method for generating and enhancing pseudo boundaries and harnessing the capabilities of LLMs to produce higher-quality coherent captions. 
Given the enhanced captions, their corresponding boundaries are generated and further optimized using a carefully designed unified metric and optimization algorithm, ultimately achieving optimal quality.
Moreover, we deploy this approach on a substantial volume of unannotated, large-scale video data, effectively bolstering the training data for the DVC task and thereby resulting in a notable performance improvement.

\begin{figure}[ht]
    \centering
    \includegraphics[width=\linewidth]{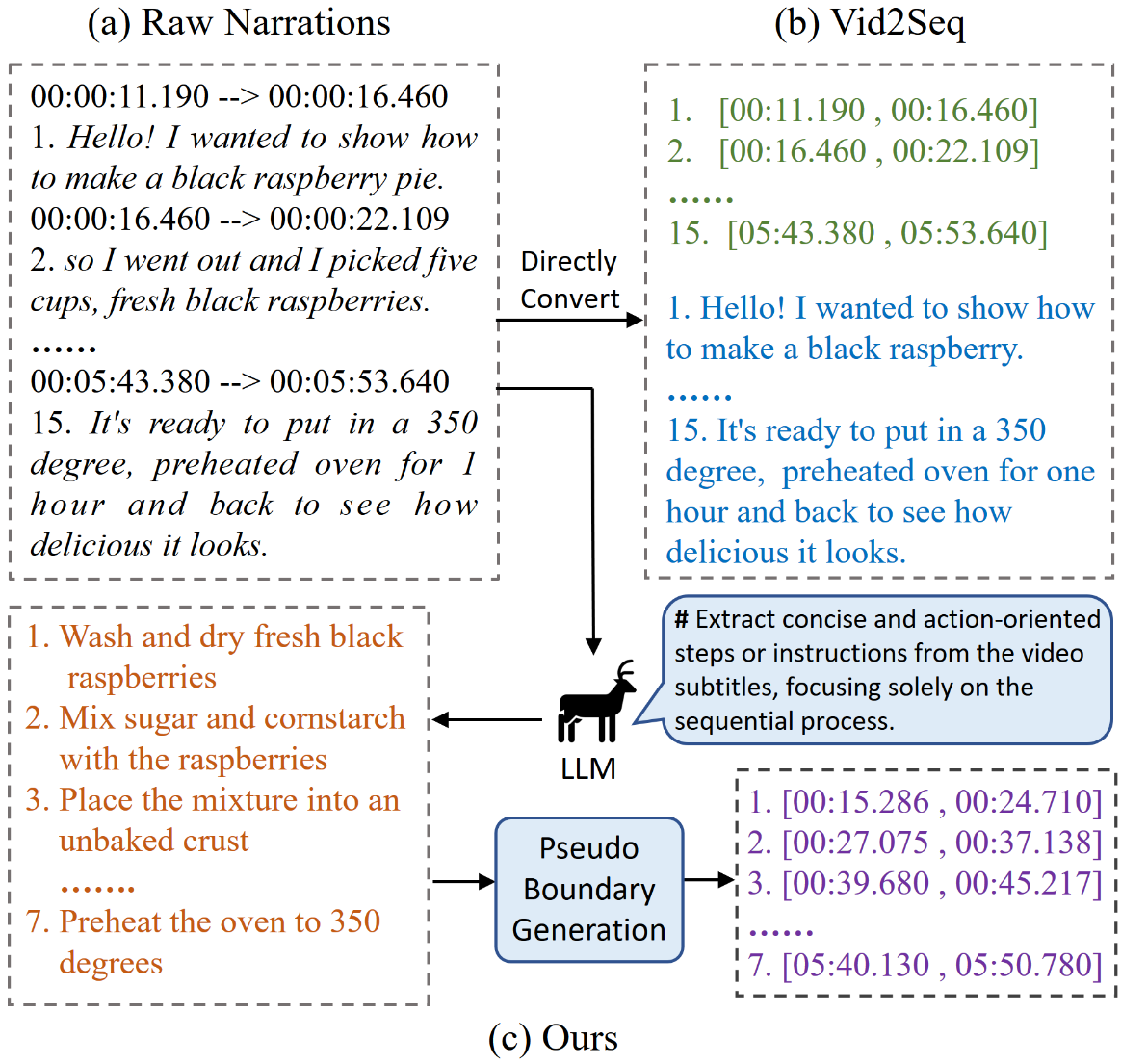}
    \caption{Comparison between Vid2Seq~\cite{Vid2Seq2023} and ours. (a) Raw subtitles extracted from videos. (b) Vid2Seq~\cite{Vid2Seq2023} directly converts the raw text and timestamps of subtitles into pseudo event boundaries and captions for pretraining. (c) Our proposed framework utilizes LLMs to generate rich and accurate captions for events from raw narrations. Subsequently, corresponding pseudo boundaries can be generated using these captions through our devised optimization algorithm (cf. Sec.~\ref{sec:pseudo_box}).}
    \label{fig:intro}
\end{figure}

Similar to our motivation, Vid2Seq~\cite{Vid2Seq2023} also emphasizes the utilization of large-scale unlabeled video data for model training to boost performance. They first propose a single-stage DVC framework that collectively predicts all event captions and corresponding temporal boundaries by generating a single sequence of discrete tokens. Then, this framework is pre-trained on numerous unlabeled narrated videos, where the target event captions and boundaries are obtained directly from the text and timestamps of subtitles (as shown in Figure~\ref{fig:intro}(b)). Nevertheless, these raw subtitles usually comprise dialogues, personal reflections, or superfluous background details, and are often temporally misaligned with the visual stream, which introduces significant noise into the event learning. Therefore, while Vid2Seq successfully integrates a larger scale of data for DVC, the effectiveness and utilization efficiency of its large-scale video data still need to be improved.

To this end, we present \textbf{DIBS}, a novel DVC framework that generates accurate event captions along with pseudo boundaries from unlabeled videos to enable effective pretraining for DVC. Specifically, DIBS harnesses diverse off-the-shelf LLMs, leveraging their proficiency in text processing to produce coherent and rich caption candidates for events from unlabeled videos. 
Moreover, it generates corresponding pseudo event boundaries for each caption, and further optimizes each boundary with a meticulously designed algorithm considering multiple metrics such as diversity, event-centricity, temporal ordering, and coherence. 
Furthermore, considering that the generated pseudo boundary is still imperfect and includes noises (e.g., background segments), we propose a novel training strategy with online boundary refinement. This strategy aims to iteratively refine and improve the pseudo boundaries during the training phase. Additionally, we seamlessly integrate our approach with state-of-the-art DVC training frameworks such as PDVC~\cite{PDVC2021}, achieving significantly improved results on extensive benchmarks.

\section{Related Work}
\subsection{Dense Video Captioning}

Dense video captioning entails event localization and captioning tasks in long-form videos. Early approaches to this problem typically employed a two-stage "detect-then-describe" framework. Within this framework, previous methods~\cite{DVC2017,SDVC2019,HCN2018,MMDEV2-2022} focused extensively on improving event representation. For instance, HCN~\cite{HCN2018} observed that contextual modeling could significantly enhance event captioning performance, while~\cite{MMDEV2020,MMDEV2-2022} incorporated the audio modality to produce a more robust representation.

Traditionally, the separation between event localization and captioning posed a gap in methods. However, recent approaches~\cite{jointDVC2018,SGR2021,E2ESG2022,Vid2Seq2023,MT2018} strive for joint learning. MT~\cite{MT2018} connected captioning loss and proposal boundaries via a differential masking mechanism for their mutual optimization. PDVC~\cite{PDVC2021}, inspired by DETR~\cite{DETR2020}, framed the task as set prediction, enabling simultaneous optimization of both tasks. In contrast, SGR~\cite{SGR2021} proposed a top-down framework, generating paragraphs before assigning event descriptions to video segments. E2ESG~\cite{E2ESG2022} tackled dense video captioning as a unified sequence-to-sequence task using a multimodal Transformer, predicting event locations and captions concurrently.


Despite their advancements, current DVC algorithms still rely on comprehensive annotations for events, particularly precise event boundaries. This requirement restricts the utilization of large-scale datasets to enhance DVC performance. Recently, Vid2Seq~\cite{Vid2Seq2023} made a groundbreaking attempt by leveraging unlabeled narrated videos for DVC pre-training. However, it directly converts the text and timestamps from narrations into pseudo event captions and boundaries for training, which introduces considerable noise (e.g., backgrounds) into the learning targets. This operation limits the potential benefits of utilizing extensive narrated video datasets for DVC. 
\subsection{Weakly-supervised Dense Video Captioning}
Research in weakly-supervised dense video captioning has gained traction due to its potential to bypass the tedious task of annotating precise event boundaries in lengthy videos. WSDEV~\cite{WSDEC2018} introduced a cyclical system tackling caption generation and sentence localization as dual tasks.  WLT~\cite{multi-modal-WSDEC2019} extended this by incorporating audio inputs for improved event captioning. Recently, EC-SL~\cite{EC-SL2021} introduced a concept learner to enhance the sentence localizer. However, these methods primarily focus on reducing reliance on precise boundary annotations without incorporating unlabeled data for training or significantly improving performance compared to fully supervised approaches.

\subsection{Video-Language Pretraining} 
 Large-scale video-text pretraining has proven highly effective in diverse video applications like video retrieval, recognition, and non-dense video captioning. UniVL~\cite{Luo2020UniVL}, for instance, employed multi-task pretraining on an instructional dataset for video retrieval. All-in-one~\cite{All-in-one2023} extended this approach with an end-to-end video-language model, incorporating multiple video datasets to support various downstream tasks like video question answering.

However, few studies have focused on extensive pretraining for dense video captioning due to the demanding nature of event annotation in these tasks. UEDVC~\cite{UEDVC2022} introduced a pretraining task on ActivityNet to enhance DVC performance on the same dataset. Vid2Seq~\cite{Vid2Seq2023} collected and used numerous narrated videos for pretraining in dense video captioning. However, it directly translates raw narrations and timestamps into pseudo event captions and boundaries for DVC pretraining, leading to significant noise and reduced effectiveness in utilizing large-scale narration video data. In contrast, our approach introduces a new pipeline to extract rich and accurate event captions and pseudo boundaries, preserving the wealth of information in large-scale videos for DVC.
 
\section{Approach}
\label{sec:formatting}
Given an input video $\textbf{V}$ comprising $M$ frames $\textbf{f}$, DVC aims to temporally localize and describe events within the video using natural language. In particular, DVC must predict timestamped boundaries $\textbf{b}$ around key moments and generate descriptive captions $\textbf{c}$ for each segmented event, resulting in a triplet 
$(\{\textbf{f}_m|m \in M\}, \{\textbf{b}_n | n \in N\}, \{\textbf{c}_n | n \in N\})$,    
where $N$ represents the number of events.

Therefore, large-scale pretraining for DVC requires an extensive collection of video data paired with textual descriptions like subtitles or speech-to-text transcriptions, represented as $(\textbf{V}, \textbf{C})$. However, this data lacks precise boundary information crucial for determining event quantity, positions, and durations. Additionally, the textual descriptions may not consistently align with the requirements of dense captioning, needing a coherent, sequentially narrated description based on pivotal events.


Previous methods, like Vid2Seq \cite{Vid2Seq2023}, tackled this issue by segmenting subtitles into boundaries and captions. However, these subtitles often contain dialogues, musings, reflections, or irrelevant details, leading to discrepancies between the intended content and the subtitles. This mismatch resulted in inaccuracies in automatically generated captions, misrepresenting the depicted events in the video.

\begin{table}[t]
\centering
\resizebox{\linewidth}{!}{
\begin{tabular}{l|c|c}
\toprule
Method & Vid2Seq~\cite{Vid2Seq2023} & Ours \\ 
\midrule
$N$ & \#Sentences in subtitle & \#Events reinterpreted by diverse LLMs\\
$\textbf{c}$ & Noisy raw sentences in subtitle & Rich, event-centric and sequential \\
$\textbf{b}$ & Timestamps of subtitle & Our pseudo boundaries generator in \cref{sec:pseudo_box} \\
\bottomrule
\end{tabular}}
\caption{Comparison of event caption generation methods between Vid2Seq~\cite{Vid2Seq2023} and our approach in terms of `$\textbf{b}$', `$\textbf{c}$', and `$N$'.}
\label{tab:compare-to-vid2seq}
\vspace{-0.5cm}
\end{table}

\subsection{Prompting LLMs for DVC-Oriented Captions}
\label{sec:llm}

To address the aforementioned issue more effectively, we aim to utilize the capabilities of off-the-shelf LLMs~\cite{touvron2023llama2,2023internlm} to enrich $\textbf{C}$ for the DVC task.
LLMs are recognized for their proficiency in text processing, showcasing a remarkable capacity to generate rich and contextually accurate captions, particularly when provided with carefully crafted prompts.


\noindent\textbf{Diverse Off-the-shelf LLMs} Our exploration into diverse LLMs includes both open-source models like LLAMA-2~\cite{touvron2023llama2} and InternLM~\cite{2023internlm}, as well as closed-source API-based models such as ChatGPT and Claude. Leveraging diverse LLMs enables us to evaluate different backbones and datasets, tapping into the distinct capabilities of each model. Tailoring carefully selected prompts to the unique characteristics of each LLM, whether open-source or API-based, enables us to extract accurate and contextually relevant event captions. Integrating a diverse range of LLMs illustrates the adaptability and effectiveness of our methodology, highlighting its versatility across different LLMs and ensuring the extraction of high-quality event captions.


\noindent\textbf{Leveraging Sparse Ground Truth Captions as Prompts for Prompt Generation} Initially, we aim to extract the event events descriptions $\{\textbf{c}_n | n \in N\}$ from the subtitles $\textbf{C}$. To achieve this, we employ a circle-prompting strategy, initially providing a small set of ground truth event captions as hints and querying an LLM for prompts that can generate similar results.
We subsequently conduct iterative testing of prompts and captions for better results and manually correct LLM errors in the loop to ensure that our prompts generate rich, accurate, concise, coherent, and event-centric captions. This iterative process seeks to achieve a balanced compromise between precision and conciseness. An example prompt we generate is specified as follows:
\\\textit{Task: Extract concise and action-oriented steps or instructions from the video subtitles, focusing solely on the sequential process. Each step should be presented as a single sentence with clear actions. Exclude any steps that are not directly related to the actions in the video. Generate the steps directly without repeating the original text.}



Consequently, the original subtitles are transformed into logically structured and temporally coherent description of events. Table~\ref{tab:compare-to-vid2seq} and Figure~\ref{fig:intro} present a comparison between our method and Vid2Seq~\cite{Vid2Seq2023} in generating event captions, accompanied by an  example. Utilizing various LLMs and prompts, we generate multiple candidate event captions. These candidates, introduced in the next section during the pseudo boundary generation process, are optimized with boundary generation.

\subsection{Optimization of Pseudo Boundaries}
\label{sec:pseudo_box}

After generating event captions $\{\textbf{c}_n | n \in N\}$, the subsequent challenge is to ascertain the temporal boundaries $\{\textbf{b}_n | n \in N\}$ of the corresponding events. 
In this section, we introduce how to obtain and optimize the corresponding boundaries for each event caption.
The optimization comprises two main objectives: first, maximizing alignment between the event caption $\textbf{c}_n$ and the video clip $\textbf{f}_{[\textbf{b}_n]}$; and second, ensuring that the temporal order relationships between the boundaries reflect those between event captions.

\noindent\textbf{Vision-Language Similarity Matrix}
For semantic alignment between $\{\textbf{c}_n | n \in N\}$ and $\{\textbf{f}_m|m \in M\}$, we adopt a bottom-up optimization strategy. Initially, a pre-trained vision-language (VL) model, denoted as $(M_V, M_L)$ for vision and language feature extractors, is employed to calculate similarities between individual frames in the video and each caption. This yields a similarity matrix $\textbf{S}$ that signifies the associations between frames and captions.
\begin{equation}
\label{similarity}
    \textbf{S}_{m,n} = \frac{ M_V(\textbf{f}_m) \cdot M_L(\textbf{c}_n) }{ | M_V(\textbf{f}_m) | \cdot | M_L(\textbf{c}_n) |}
\end{equation}
\emph{Note that when employing a video-language model, a short video clip centered around a frame is utilized to represent this frame.}
For implementation, image-language models such as CLIP~\cite{radford2021clip} and video-language models like UniVL~\cite{Luo2020UniVL} are utilized. In the experiments, we aggregate scores from multiple vision-language models, averaging them to yield a more robust similarity matrix. 

\emph{Discussion: the domain gap and noisy detection issue.} Evidently, the quality of the similarity matrix significantly depends on the VL models' quality and the domain gap between the dataset used for VL model training and the DVC dataset. On the one hand, this underscores the importance of employing diverse LLMs to generate rich caption candidates. On the other hand, despite these concerted efforts, the similarity matrix still exhibits notable detection noise and, in some instances, false positives.




\noindent\textbf{Caption-Aware Pseudo Boundary Generation with Soft Time Constraints} Leveraging the similarity matrix $\textbf{S}$, 
we convert the event localization problem into the identification of the optimal $N$ frames in $\{\textbf{f}_m|m \in M\}$ corresponding to captions $\{\textbf{c}_n | n \in N\}$. This process should adhere to the temporal order specified in $\{\textbf{c}_n | n \in N\}$, resulting in a drop dynamic time warping (Drop-DTW)~\cite{dvornik2021dropdtw} problem. We denote this baseline method as the \textbf{Drop-DTW Baseline}.


However, due to the noisy detection issue discussed above, the effectiveness of the direct Drop-DTW Baseline method is constrained. This ineffectiveness arises from several reasons: 1) \textbf{Multimodal Responses:} Higher response positions often exhibit multiple peaks, with their distribution in the video lacking concentration. 2) \textbf{Non-Uniform Event Coverage:} Events do not exhibit a preference for a uniformly distributed coverage across the majority of the entire video. 3) \textbf{Strict Temporal Ordering Constraints:} the hard global temporal ordering constraints in Drop-DTW not only slow down the optimization process significantly but also impede individual captions from effectively finding the best-matching video intervals; 4) \textbf{Duration Determination:} Additionally, determining the duration of events lacks a straightforward, intuitive method.


\begin{algorithm}[t]
\caption{Pseudo Boundary Generation}
\label{alg:pseudo_box_generation}
\small
\begin{algorithmic}[1]
\State \textbf{Input:} event captions $\{\textbf{c}_n | n \in N\}$; similarity matrix $\textbf{S}$; user-defined top-$k$ parameter $K$, total iterations $Q$.
\State \textbf{Initialization:} $\{\textbf{b}^0_n | n \in N\} \gets $ Divide the entire video $\textbf{V}$ into $N$ equal segments.
\For {each caption $\textbf{c}_n$ in $\{\textbf{c}_n | n \in N\}$}
    \For {each iteration $t$ in $t \in Q$}
        \State collect top-$\hat{k}$ frames $\{T_{\hat{k},n}|\hat{k} \in K\}$ around $\textbf{b}^{t}_n$.
        \For {each $\hat{k}$ in $K$}
            \State $D_{\hat{k},n} = \sum_{i=1}^{K}|T_{i,n} - T_{\hat{k},n}|$
        \EndFor
        \State $\hat{k}^* \gets \hat{k} \text{  with minimum  } D_{\hat{k},n}.$
        \State $ \text{new boundary center}  \gets T_{\hat{k}*,n}$
        \State $ \text{new boundary size} \gets 2*\alpha*std_n.$ 
        \State update boundary $ \gets \textbf{b}_n^t $ in Eq.~\ref{finebox} 
        \State record loss $ \gets L^t $  in Eq.~\ref{loss}.
    \EndFor
    \State select iter with minimum loss $q = \underset{t}{\mathrm{argmin}}\, L^t, t \in Q$
    \State determine the pseudo boundary for caption $\textbf{c}_n$: $\textbf{b}_n \gets \textbf{b}_n^{q}$.
    \State determine the loss for captions : $L \gets L^{q}$.
\EndFor
\State \textbf{Output:} Generated pseudo boundaries $\{\textbf{b}_n | n \in N\}$ and loss $L$ for input event captions $\{\textbf{c}_n | n \in N\}$.
\end{algorithmic}
\end{algorithm}

\begin{figure}[t]
    \centering
    \includegraphics[width=0.9\linewidth]{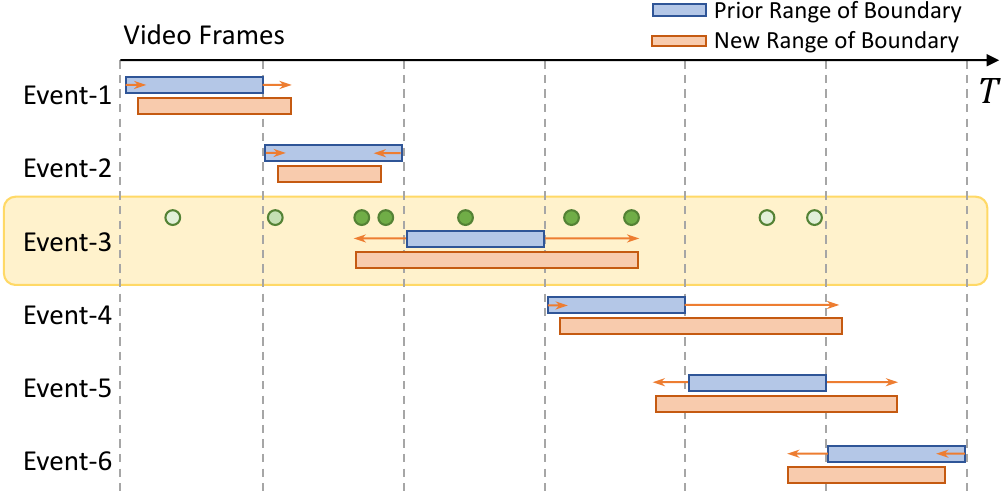}
    \caption{Prior range division and frame processing in pseudo boundary generation. Frames showcasing top-$k$ global similarity to Event-3 are indicated by deep \green{green} points, contrasting with lighter green points denoting similarity outside the top-$k$ range. The \blue{blue} box delineates the evenly divided prior range, while the \orange{orange} box signifies the adjusted range after an iteration. The pseudo boundary generation process is outlined in \cref{alg:pseudo_box_generation}.}
    \label{fig:order}
    \vspace{-0.7cm}
\end{figure}

To address these issues, we propose a caption-aware pseudo boundary generation algorithm with soft time constraints, as illustrated in Algorithm~\ref{alg:pseudo_box_generation}. 
To encourage events cover most of the video length \emph{evenly}, we initialize the boundaries $\textbf{b}^0$ by dividing the video into $N$ equal segments. Subsequently, we iteratively optimize the boundaries, as illustrated in Figure~\ref{fig:order}. 
For the $n$th caption, in its $t$th iteration, we first collect a \emph{local} set of top-$\hat{k}$ frames around $\textbf{b}^{t-1}_n$ (the boundary range in the previous iteration, starting from $\textbf{b}^{0}_n$) with the highest similarity scores to $\textbf{c}_n$. From these $\hat{k}$ positions, we select the new boundary center. For each of the $\hat{k}$ frames, we calculate the total temporal distance from this frame to the other $\hat{k}-1$ frames. The frame with the minimum total distance indicates that these top $\hat{k}$ frames are more concentrated around it, thus setting it as the new boundary center, demoted as $T_{\hat{k}^*,n}$. 
The size of the boundary is determined by calculating the standard deviation with respect to the new boundary center as shown in Equation~\ref{std}.  
\begin{equation}
\label{std}
    std_n = (\frac{1}{k}\sum_k(T_{k,n}-T_{\hat{k}^*,n})^2)^{\frac{1}{2}} 
\end{equation}
To this end, we gain a coarse boundary $\textbf{b}_{coarse,n}$ and we further determine the actual boundary $\textbf{b}_n^t$ of the boundary by selecting the minimum and maximum values of $k$ frames within the new coarse boundary as  depicted in Equation~\ref{finebox}:
\begin{equation}
\label{finebox}
\begin{aligned}
    \textbf{b}_{coarse,n}=[T_{\hat{k}^*,n}-\alpha*std_n, T_{\hat{k}^*,n}+\alpha*std_n] \\
\textbf{b}_n^t = [\min(T_{k,n}), \max(T_{k,n}) \text{ for } T_{k,n} \text{ in } \textbf{b}_{coarse,n}]
\end{aligned}
\end{equation}
where $\alpha$ is a hyperparameter determining the size of the coarse boundary. After each iteration, we define a loss function to evaluate the quality of the generated boundary, as shown in Equation~\ref{loss}: 
\begin{equation}
\label{loss}
\begin{aligned}
    L^t &= \sum_{n}\sum_{k} \mathbf{S}_{k,n} \cdot \textit{Dis}(T_{k,n}, \mathbf{b}_n^t), \\
    \textit{Dis}(f,\mathbf{b}) &= 
    \begin{cases}
        -\min(f -\mathbf{b}_s, \mathbf{b}_e - f), & \text{if } \mathbf{b}_s < f < \mathbf{b}_e, \\
        \max(\mathbf{b}_s - f, f - \mathbf{b}_e), & \text{if } f < \mathbf{b}_s \text{ or } f > \mathbf{b}_e.
    \end{cases}
\end{aligned}
\end{equation}
Here, the loss is the total distance weighted by the similarity score between the frame and caption. The function $\textit{Dis}(\cdot)$ measures the distance between a frame $f$ and the boundary $\textbf{b}$, where $\textbf{b}_s$ and $\textbf{b}_e$ represent the start and end of the boundary respectively. The function $\textit{Dis}(f, \textbf{b})$ is positive when $f$ is within the boundary range and negative otherwise.   We then collect the local top-$k$ frames with the highest similarity scores to the $n$th caption around the new boundary and proceed to the next iteration. Finally, we select the boundary with the minimum loss value as the final optimized boundary. Refer to Algorithm~\ref{alg:pseudo_box_generation} for further details on generating the pseudo boundaries.

\subsection{Training with Online Boundary Refinement}


Although we can employ various fully supervised approaches~\cite{PDVC2021,UEDVC2022,E2ESG2022} to train a DVC model using generated event captions and pseudo boundaries, these boundaries might be imperfect, encompassing inaccuracies like background elements. Utilizing them directly as ground truth during DVC pretraining could mislead the event learning process, especially in event localization. To counter this, we introduce an online strategy for refining pseudo boundaries, seamlessly integrating it with cutting-edge DVC training frameworks like PDVC~\cite{PDVC2021}.

\noindent
\textbf{Background Review of PDVC} 
PDVC~\cite{PDVC2021} is an effective DVC framework employing parallel decoding and set prediction principles. It starts by utilizing a pre-trained video feature extractor and a transformer encoder to derive a sequence of frame-level features. Then, employing $M$ learnable event queries ${\{\mathbf{q}_i\}}_{i=1}^M$, it employs a transformer decoder along with three prediction heads (localization head, caption head, and event counter) to simultaneously predict $M$ boundaries, $M$ captions, and event count. During inference, the model selects the top detected events by ranking captioning and localization scores, without using non-maximum suppression (NMS)~\cite{RCNN2015}.

\begin{figure}
    \centering
    \includegraphics[width=0.48\textwidth]{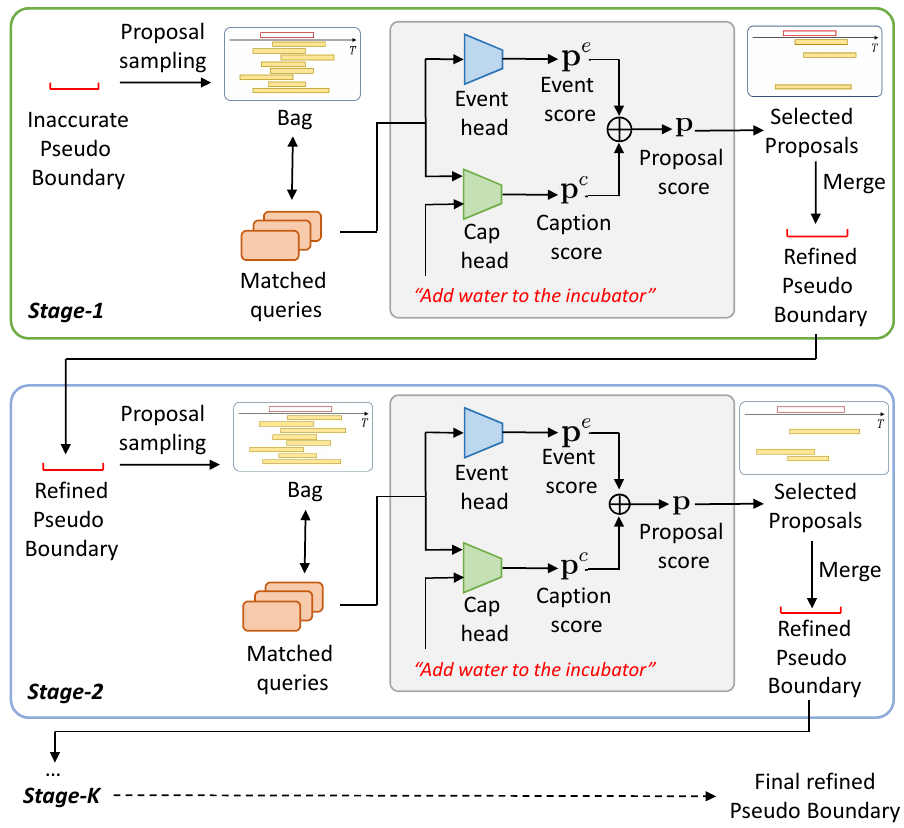}
    \caption{The process of online pseudo boundary refinement.}
    \label{fig:refinement}
    \vspace{-0.5cm}
\end{figure}

\noindent
\textbf{Online Pseudo Boundary Refinement} 
We refine pseudo boundaries by augmenting their quantity and conducting quality evaluations. 
As shown in Figure~\ref{fig:refinement}, consider a pseudo boundary $\textbf{b} = (t,d)$ of caption $\mathbf{c}$, we employ a standard jitter augmentation on the boundary duration $d$ and the boundary center $t$ with jitter ratio $r_1$ and $r_2$ respectively, obtaining an augmented set of proposals, denoted as $\mathcal{B} = \{\hat{\textbf{b}}_u\}_{u=1}^U$
($U$ proposals in total and $\textbf{b}$ is contained in $\mathcal{B}$ as well), representing potentially superior event segments for caption $\mathbf{c}$. Then, following PDVC, we adopt the Hungarian matching method~\cite{DETR2020} between all proposals $\{\hat{\textbf{b}}_u\}_{u=1}^U$ and query embeddings $\{\mathbf{q}_i\}_{i=1}^M$ to link each proposal with a specific query, yielding $\{\textbf{Q}_u\}_{u=1}^U$ ($\textbf{Q}_u $ $=$ $\mathbf{q}_i$, if the $i$th query is linked to the $u$th proposal). The linked query serves as the proxy feature for the proposal.

Utilizing the augmented set of proposals and their proxy features, we evaluate their quality online during training. PDVC offers an event classification head $\mathbf{h}^e$ and a caption scoring head $\mathbf{h}^c$, both using the query embeddings as input and producing confidence scores. Formally,
\begin{gather}
    \mathbf{p}^e_u = \texttt{Sigmoid}(\mathbf{h}^e(\mathbf{Q}_u)) \\
    \mathbf{p}^c_u = \texttt{Softmax}_{\{U\}}(\mathbf{h}^c(\mathbf{Q}_u, \mathbf{c}))
\end{gather}
Note that the event score $\mathbf{p}^e_u$ is normalized with the Sigmoid activation while the caption score $\mathbf{p}^c_u$ is normalized with the Softmax activation across all proposals in $\mathcal{B}$. The final assessment score $\mathbf{p}_u$ for the proposal $\hat{\mathbf{b}}_u$ is calculated as $\mathbf{p}_u = \mathbf{p}_u^e + \mathbf{p}_u^c$. A higher proposal score means this proposal better represents the boundary of caption $\textbf{c}$.


Finally, we select the top-$K$ proposals in $\{\hat{\textbf{b}}_u\}_{u=1}^U$ with the highest scores and compute a weighted average of their boundaries as the final refined boundary.
\begin{equation}
    \textbf{b}_{ref} = \dfrac{\sum_{k=1}^K \mathbf{p}_k \cdot \hat{\mathbf{b}}_k}{\sum_{k=1}^K \mathbf{p}_k}
\end{equation}
The refined pseudo boundary $\textbf{b}_{ref}$ replaces the original boundary $\textbf{b}$ for subsequent training stages. This iterative process, shown in Figure~\ref{fig:refinement}, refines each pseudo boundary in multiple stages, resulting in increasingly accurate boundaries. Our implementation defaults to a 2-stage refinement process to derive the final boundary. Importantly, boundary refinement doesn't impact model inference, incurring no additional computational overhead during this phase.

\section{Experiments}

\subsection{Experimental Setting}
\paragraph{Datasets} Our experiments encompass two prominent datasets commonly employed for dense video captioning: YouCook2~\cite{ZhXuCo18youcook2} and ActivityNet Captions~\cite{krishna2017anet}. For pretraining, we leverage a subset of the HowTo100M~\cite{miech19howto100m} dataset, specifically focusing on cooking videos, amounting to approximately 56,000 videos.

\paragraph{Implementation Details}
In our setup, we uniformly sample video frames at 1 FPS and adjust them to a fixed count denoted as $F$. For YouCook2, $F$ is set to 200, and for ActivityNet Captions, it's 100. We utilize pretrained vision-language models like CLIP~\cite{radford2021clip} and UniVL~\cite{Luo2020UniVL} to extract frame-level features across all datasets. The model undergoes a two-stage training process: initially, a 10-epoch pretraining on a subset of HowTo100M, followed by a 20-epoch fine-tuning phase on each target dataset. Notably, to address the domain gap between the HowTo100M dataset and the target dataset, we augment the training data by incorporating the target dataset using pseudo-boundaries during pretraining. Our model architecture mirrors PDVC~\cite{PDVC2021}, incorporating a transformer encoder, transformer decoder, and three prediction heads.



\paragraph{Evaluation Metrics} We employ standard captioning metrics: METEOR~\cite{banerjee2005meteor} (M) for semantic similarity, and CIDEr~\cite{vedantam2015cider} (C) for human judgment correlation. For event localization evaluation, we utilize average precision (Pre.) and recall (Rec.) at Intersection over Union(IoU) thresholds of 0.3, 0.5, 0.7 and 0.9, along with the overall F1 score for boundary prediction. These metrics are computed using the official ActivityNet challenge toolbox. Additionally, we incorporate the SODA\_c \cite{fujita2020soda} (S) metric, which provides a joint evaluation of caption quality and localization accuracy. 

\begin{table*}[t]
\centering
\resizebox{\linewidth}{!}{
\begin{minipage}[t]{0.55\textwidth}
\centering
\resizebox{\textwidth}{!}{%
\begin{tabular}{c|l|cc|ccc|ccc}
\toprule
\multirow{2}{*}{Training} & \multirow{2}{*}{Method} & \multicolumn{2}{c|}{Settings} & \multicolumn{3}{c|}{YouCook2} & \multicolumn{3}{c}{ActivityNet} \\&  & Pretrain & Backbone & M & C & S & M & C & S \\
\midrule
\multirow{5}{*}{\makecell{Weakly- \\Supervised}} 
& WLT~\cite{multi-modal-WSDEC2019} & $\emptyset{}$ & TSN
& - & - & -
& 4.93 & 13.79 & - \\
& WS-DEC~\cite{WSDEC2018}$^\dag$  & $\emptyset{}$ & C3D
& - & - & - 
& 6.30 & 18.77 & - \\
& EC-SL~\cite{EC-SL2021}$^\dag$  & $\emptyset{}$ & C3D 
& - & - & -
& 7.49 & 21.21 & - \\
& DIBS (Ours) & $\emptyset{}$ & CLIP 
& 4.63 & 22.05 & 4.03
& 7.33 & 15.86 & 4.67 \\ 
& DIBS (Ours)  & $\emptyset{}$ & UniVL
& 5.90 & 29.62 & 4.96
& 6.76 & 13.75 & 4.26 \\ 
\midrule
\multirow{8}{*}{\makecell{Fully- \\Supervised}}
& PDVC~\cite{PDVC2021}  & $\emptyset{}$ & TSN
& 4.74 & 22.71 & 4.42
& 7.96 & 28.96 & 5.44 \\
& PDVC~\cite{PDVC2021}$^\dag$   & $\emptyset{}$ & CLIP
& 5.47 & 28.37 & 5.00 
& 8.31 & 30.11 & 5.63 \\
& PDVC~\cite{PDVC2021}$^\dag$  & $\emptyset{}$ & UniVL
& 7.87 & 46.02 & 6.87 
& 8.24 & 28.21 & 5.43 \\
& UEDVC~\cite{UEDVC2022} & 676k & TSN 
& 2.18 & 8.37 & 3.34
& - & - & 5.49 \\
& E2ESG~\cite{E2ESG2022}  & $\emptyset{}$ & C3D
& 3.49 & 25.00 & -
& - & - & - \\
& Vid2Seq~\cite{Vid2Seq2023}  & 15M & CLIP 
& 9.30 & 47.10 & 7.90 
& 8.50 & 30.10 & 5.80 \\
& \textbf{DIBS (Ours)} & 56k & CLIP
& 7.51 & 44.44 & 6.39 
& \textbf{8.93} & \textbf{31.89} & \textbf{5.85} \\
& \textbf{DIBS (Ours)} & 56k & UniVL
& \textbf{9.41} & \textbf{59.35} & \textbf{7.97}
& 8.25 & 28.85 & 5.35 \\
\bottomrule
\end{tabular}}
\vspace{-3pt}
\caption{\small Performance of caption generation on YouCook2 and ActivityNet. $^\dag$ Results are obtained from our implementation with official codebase.}
\label{tab:caption_compare}
\end{minipage}
\hspace{4pt}
\begin{minipage}[t]{0.45\textwidth}
\centering
\resizebox{\textwidth}{!}{%
\begin{tabular}{l|cc|cc|cc}
\toprule
\multirow{2}{*}{Method} & \multicolumn{2}{c|}{Settings} & \multicolumn{2}{c|}{YouCook2} & \multicolumn{2}{c}{ActivityNet} \\  & Pretrain & Backbone & Rec. & Pre. & Rec. & Pre. \\
\midrule
 PDVC~\cite{PDVC2021}  & $\emptyset{}$ & TSN
& - & - 
& 55.42 & 58.07  \\
 PDVC~\cite{PDVC2021}$^\dag$   & $\emptyset{}$ & CLIP
& 21.76 & 31.92 
& 50.82 & 55.73  \\
 PDVC~\cite{PDVC2021}$^\dag$  & $\emptyset{}$ & UniVL
& 29.66 & 42.04 
& 53.21 & 59.46  \\
 UEDVC~\cite{UEDVC2022} & 676k & TSN 
& - & - 
& \textbf{59.00} & \textbf{60.32}  \\
 E2ESG~\cite{E2ESG2022}  & $\emptyset{}$ & C3D
& 3.49 & 25.00 
& - & -  \\
 Vid2Seq~\cite{Vid2Seq2023}  & 15M & CLIP 
& 27.90 & 27.80 
& 52.70 & 53.90  \\
 \textbf{DIBS (Ours)} & 56k & CLIP
& 26.24 & 39.18 
& 53.14 & 58.31   \\
 \textbf{DIBS (Ours)} & 56k & UniVL
& \textbf{30.80} & \textbf{45.13} 
& 53.02 & 58.39 \\
\bottomrule
\end{tabular}}
\vspace{4pt}
\caption{\small Performance of event localization on YouCook2 and ActivityNet. $^\dag$ Results are obtained from our implementation with official codebase.}
\label{tab:localization_compare}
\end{minipage}}
\vspace{-0.5cm}
\end{table*}

\subsection{Comparison with State-of-the-art Methods}
In Table~\ref{tab:caption_compare}, we compare our captioning performance against state-of-the-art approaches. We evaluate our model under two settings: weakly supervised and fully supervised paradigms. In the weakly supervised setting, our model is trained directly on the target dataset using ground truth captions and generated pseudo boundaries, without using any ground truth boundaries. It can be observed that this strategy achieves comparable performance compared to prior weakly supervised DVC methods, demonstrating the effectiveness of our pseudo boundary generation and refinement.
Regarding the standard fully-supervised setting, our approach outperforms PDVC~\cite{PDVC2021} with the same backbone architecture after pretraining. This illustrates the significant benefits of pretraining for DVC. Particularly, on YouCook2 and ActivityNet, our models with UniVL and CLIP backbones surpass previous methods across multiple metrics. Notably, Vid2Seq incorporates additional speech cues during inference, providing a richer multi-modal context. Besides, Vid2Seq uses much larger-scale pretraining data (15M videos) and captioning models (T5~\cite{T52019}) compared to our DIBS. These factors likely explain some performance differences relative to Vid2Seq, particularly with the CLIP backbone. However, using the UniVL backbone, our model can surpass Vid2Seq on YouCook2 by a significant margin. On ActivityNet, our CLIP-based approach also achieves state-of-the-art results.

In Table~\ref{tab:localization_compare}, we see that our pretraining consistently enhances the event localization, enabling our DIBS to outperform previous methods on YouCook2 by a great margin. However, on ActivityNet, our approach lags behind PDVC and UEDVC~\cite{UEDVC2022} while outperforming Vid2Seq. This disparity may be due to the domain gap between instructional and activity videos, impacting the localization of less common events despite improvements in captioning.

In summary, our proposed pretraining approach substantially boosts event localization on well-matched instructional datasets like YouCook2.  Regarding general activity localization, additional techniques may be necessary to transfer specific knowledge from the instructional videos due to the clear domain gap.  Therefore, taking careful consideration of domain similarity is critical when leveraging unlabeled videos for DVC pretraining.

\subsection{Ablation Study}
\paragraph{Effects of Pseudo Boundary}
We examine the impact of pseudo boundaries on YouCook2, comparing against a baseline where ground truth event boundaries are omitted. When utilizing pseudo boundaries, matching between proposals and captions primarily relies on the Generalized IoU (GIoU) cost. To provide a comprehensive comparison, we construct two additional settings, both incorporating caption information similar to our pseudo boundary. The first setting introduces an additional caption cost for matching, replacing the GIoU cost, while the second setting incorporates a caption-proposal similarity cost in place of the GIoU cost. In Table~\ref{table:pbox}, we present a comprehensive comparison of model performance on event localization and caption generation using CLIP and UniVL as backbones. The results demonstrate a significant improvement in model performance when employing pseudo boundaries, whereas the other two settings exhibit inferior performance compared to the baseline at most metrics. This corroborates the effectiveness of generating pseudo boundaries using captions as a powerful method for leveraging caption information in event localization tasks. 

\begin{table}[tb]
\centering
\vspace{-0pt}
\begin{center}
\setlength\tabcolsep{6pt}
\resizebox{1.\linewidth}{!}{
\begin{tabular}{cccc|ccccc}
\toprule
& \multicolumn{3}{c|}{Settings} & \multicolumn{5}{c}{Metrics} \\ & Feature & Boundary & Caption & \small{M} & \small{C} & \small{S} & \small{Rec.} & \small{Pre.} \\
\midrule
1. & CLIP & \xmark & \xmark
& 2.32 & 9.07 & 3.40
& 18.47 & 25.01 \\
2. & CLIP &\xmark & Cost
& 1.68 & 7.09 & 2.79
& 14.19 & 16.63 \\
3. & CLIP &\xmark & Sim
& 2.18 & 8.37 & 3.34
& \textbf{20.77} & 24.70 \\
4. & CLIP & \cmark & \xmark
& \textbf{4.62} & \textbf{21.93} & \textbf{3.73}
& 15.48 & \textbf{28.82} \\
\midrule
1. & UniVL & \xmark & \xmark
& 2.66 & 9.97 & 3.21
& 15.62 & 25.18 \\
2. & UniVL &\xmark & Cost
& 1.99 & 8.49 & 3.15
& 18.75 & 21.08 \\
3. & UniVL &\xmark & Sim
& 2.27 & 9.16 & 3.16
& 15.83 & 24.88 \\
4. & UniVL &\cmark & \xmark
& \textbf{5.88} & \textbf{28.04} & \textbf{4.47}
& \textbf{19.72} & \textbf{35.43} \\

\bottomrule
\end{tabular}}
\caption{\small Performance comparison of event localization and caption generation on YouCook2 with and without pseudo boundaries.}
\vspace{-0.5cm}
\label{table:pbox}
\end{center}
\end{table}


\paragraph{Effects of Soft Time Constraints}
We assess the impact of soft time constraints on pseudo boundaries by excluding ground truth boundaries from both the YouCook2 and ActivityNet. Our method without soft time constraints performs global video iteration for each caption. Table~\ref{tab:or} delineates model performance with and without soft time constraints, employing CLIP and UniVL backbones. Introducing soft time constraints notably enhances model performance in event localization and caption generation. This improvement underscores the efficacy of soft time constraints in generating higher-quality pseudo boundaries. The sequential nature of events in videos highlights the significance of maintaining temporal order, even in scenarios with potential event overlap, as observed in ActivityNet.
\begin{table}[tb]
    \centering
    \resizebox{0.48\textwidth}{!}{
    \begin{tabular}{c|cc|ccccc}
    \toprule
        Dataset &  Features & STC & M & C & S & Rec. & Pre. \\
        \midrule
        \multirow{4}{*}{YouCook2} & CLIP &\xmark &  3.92 & 17.05 & 3.57 & 14.29 & 25.87 \\
                                  & CLIP &\cmark &  \textbf{4.62} & \textbf{21.93} & \textbf{3.73} & \textbf{15.48} & \textbf{28.82} \\
        \cmidrule{2-8}
                                  & UniVL &\xmark &  5.43 & 25.57 & 4.47 & 18.05 & 32.83 \\
                                  & UniVL &\cmark &  \textbf{5.88} & \textbf{28.04} & \textbf{4.47} & \textbf{19.72} & \textbf{35.43} \\
        \midrule
        \multirow{4}{*}{ActivityNet} & CLIP &\xmark &  6.60 & 15.23 & 3.87 & 35.31 & 57.68 \\
                                  & CLIP &\cmark &   \textbf{7.23} & \textbf{16.63} & \textbf{4.68} & \textbf{42.44} & \textbf{65.46} \\
        \cmidrule{2-8}
                                  & UniVL &\xmark &  5.88 & 13.60 & 3.68 & 35.72 & 55.14 \\
                                  & UniVL &\cmark &  \textbf{6.75} & \textbf{14.13} & \textbf{4.21} & \textbf{42.24} & \textbf{64.75} \\
    \bottomrule
    \end{tabular}}
    \caption{Comparison of model performance with and without soft time constraints. STC denotes soft time constraints.}
    \label{tab:or}
\end{table}

\paragraph{Effects of Boundary Refinement}
To assess the impact of boundary refinement, we employ two distinct settings on the YouCook2 and ActivityNet Caption datasets. In the first setting, we examine performance by omitting the ground truth boundary, utilizing pseudo boundaries both with and without refinement. In the second setting, we perform pretraining on the HowTo100M subset with and without refinement, followed by fine-tuning the pretrained models on target datasets like YouCook2 and ActivityNet within our study. In Table~\ref{tab:refine}, we present a comprehensive model performance comparison with and without refinement. Refinement enhances model performance on localization and caption metrics across both YouCook2 and ActivityNet datasets. However, while most metrics improve, a few experience a slight drop, and this discrepancy varies between the YouCook2 and ActivityNet datasets. We attribute this observation to the domain gap between the two datasets.
\begin{table}[tb]
    \centering
    \resizebox{0.48\textwidth}{!}{
    \begin{tabular}{c|cc|ccccc}
    \toprule
        Dataset & Pretrain & Refine & M & C & S & Rec. & Pre. \\
        \midrule
        \multirow{4}{*}{YouCook2} & \xmark & \xmark & 5.88 & 28.04 & 4.47 & 19.72 & 35.43 \\
                                  & \xmark & \cmark & \textbf{5.90} & \textbf{29.62} & \textbf{4.96} & \textbf{30.80} & \textbf{45.13} \\
                                  \cmidrule{2-8}
                                  & \cmark & \xmark & 9.04 & 54.98 & \textbf{8.13} & \textbf{31.55} & 44.95 \\
                                  & \cmark & \cmark & \textbf{9.41} & \textbf{59.35} & 7.97 & 30.80 & \textbf{45.13} \\
        \midrule
        \multirow{4}{*}{ActivityNet} & \xmark & \xmark & 7.23 & \textbf{16.63} & \textbf{4.68} & \textbf{42.44} & 65.46 \\
                                  & \xmark & \cmark & \textbf{7.33} & 15.86 & 4.67 & 42.21 & \textbf{65.79} \\
                                  \cmidrule{2-8}
                                  & \cmark & \xmark & \textbf{8.94} & 30.49 & 5.39 & 51.45 & 57.49 \\
                                  & \cmark & \cmark & 8.93 & \textbf{31.89} & \textbf{5.85} & \textbf{53.14} & \textbf{58.31} \\
    \bottomrule
    \end{tabular}}
    \caption{Comparison between models with and without pseudo boundary refinement.}
    \label{tab:refine}
    \vspace{-0.2cm}
\end{table}




\paragraph{Effect of Pretraining}
To examine pretraining effects, we conducted a comprehensive comparison of model performance with and without pretraining using CLIP and UniVL backbones on YouCook2 and ActivityNet datasets. Table~\ref{tab:pretrain} details performance for event localization and caption generation. It indicates improved caption generation on both datasets after pretraining, yet stable improvements in event localization are not consistently observed, especially on ActivityNet. The enrichment of pseudo captions from large-scale unlabeled videos, generated by LLMs, likely contributes to enhanced caption performance. Despite employing diverse strategies to improve the quality of pseudo boundaries, the substantial distance from ground truth boundaries persists, possibly explaining the limited impact of pretraining on event localization. We also examined the impact of varying pretraining data amounts, please refer to our supplementary materials.


\begin{table}[tb]
    \centering
    \resizebox{0.48\textwidth}{!}{
    \begin{tabular}{c|cc|ccccc}
    \toprule
        Dataset & Pretrain & Features & M & C & S & Rec. & Pre. \\
        \midrule
        \multirow{4}{*}{YouCook2} & \xmark & CLIP & 5.47 & 28.37 & 5.00 & 21.76 & 31.92 \\                                  
                                  & \cmark & CLIP & 7.51 & 44.44 & 6.39 & 26.24 & 39.18 \\
                                  & \xmark & UniVL & 7.87 & 46.02 & 6.87 & 29.66 & 42.04 \\
                                  & \cmark & UniVL & \textbf{9.41} & \textbf{59.35} & \textbf{7.97} & \textbf{30.80} & \textbf{45.13} \\
        \midrule
        \multirow{4}{*}{ActivityNet} & \xmark & CLIP & 8.31 & 30.11 & 5.63 & 50.82 & 55.73 \\                                  
                                  & \cmark & CLIP & \textbf{8.93} & \textbf{31.89} & \textbf{5.85} & 53.14 & 58.31 \\
                                  & \xmark & UniVL & 8.24 & 28.21 & 5.43 & \textbf{53.21} & \textbf{59.46} \\
                                  & \cmark & UniVL & 8.25 & 28.85 & 5.35 & 53.02 & 58.39 \\
    \bottomrule
    \end{tabular}}
    \caption{Comparative analysis of model performance with and without pretraining on YouCook2 and ActivityNet Datasets.}
    \label{tab:pretrain}
\end{table}

\begin{table}[t]
    \centering
    \resizebox{0.48\textwidth}{!}{
    \begin{tabular}{c|cc|cccccc}
    \toprule
        Dataset & Pretrain & FT Data  &  M & C & S & Rec. & Pre. & F1\\
        \midrule
        \multirow{5}{*}{YouCook2} & \xmark & 100\% & 7.87 & 46.02 & 6.87 & 29.66 & 42.04 & 34.78 \\                    
                                  & \cmark & 25\% & 7.81 & 46.69 & 7.17 & 28.93 & 41.08 & 33.95 \\
                                  & \cmark & 50\% & 8.60 & 55.73 & 7.84 & 30.14 & 42.73 & 35.34 \\
                                  & \cmark & 75\%  & 9.11 & 59.09 & 7.86 & 29.97 & 44.54 & 35.83 \\
                                  & \cmark & 100\% & \textbf{9.41} & \textbf{59.35} & \textbf{7.97} & \textbf{30.80} & \textbf{45.13} & \textbf{36.61} \\
        \midrule
        \multirow{5}{*}{ActivityNet} & \xmark & 100\% & 8.31 & 30.11 & 5.63 & 50.82 & 55.73 & 53.16 \\                    
                                  & \cmark & 25\% & 8.66 & 28.24 & 5.01 & 51.08 & 56.24 & 53.54 \\
                                  & \cmark &  50\% & 8.56 & 30.09 & 5.39 & 51.96 & 56.26 & 54.02 \\
                                  & \cmark & 75\% & 8.83 & 30.80 & 5.55 & 52.90 & 57.19 & 54.96 \\
                                  & \cmark & 100\% & \textbf{8.93} & \textbf{31.89} & \textbf{5.85} & \textbf{53.14} & \textbf{58.31} & \textbf{55.61} \\
    \bottomrule
    \end{tabular}}
    \caption{Few-shot performance fine-tuning on partial data. ``FT data'' represents the percentage of data used for fine-tuning.}
    \label{tab:ftdatasize}
    \vspace{-0.5cm}
\end{table}

\paragraph{Few-shot Dense Video Captioning}
In Table~\ref{tab:ftdatasize}, we depict the model's performance concerning varying proportions of fine-tuning data on YouCook2 and ActivityNet datasets. The results show a direct association with the amount of fine-tuning data used. Notably, even with a fraction of the complete training set post-pretraining, the model surpasses the performance of the setting without pretraining and using the entire training set. This distinction is particularly prominent on YouCook2, where achieving superior performance with only half of the target training set is feasible. 
Conversely, ActivityNet requires more training data to achieve comparable advancements.



\paragraph{Backbone Influence in Dense Video Captioning}
Table~\ref{tab:pretrain} compares CLIP and UniVL features, both with and without pretraining. UniVL demonstrates better performance on YouCook2, possibly due to its pretraining on instructional videos, narrowing the domain gap. In contrast, CLIP excels on ActivityNet, leveraging its strong generalization abilities. Our findings emphasize the importance of selecting backbones pretrained on tasks aligned with the dataset. UniVL suits instructional videos like YouCook2, while CLIP's broader generalization is advantageous for diverse datasets like ActivityNet, highlighting the need for a tailored approach based on dataset characteristics and backbone pretraining specifics in dense video captioning tasks.

\section{Conclusion}
We introduce \textbf{DIBS}, a novel pretraining framework for DVC that improves the quality of pseudo event boundaries and captions derived from large-scale unlabeled videos. Leveraging LLMs, we generate and optimize pseudo boundaries and captions simultaneously, emphasizing objectives like diversity and coherence. We also propose an online boundary refinement strategy to further improve the quality of pseudo boundaries. Extensive experiments validate the effectiveness of DIBS. 

\noindent \textbf{Acknowledgments}
This work is partially supported by the National Key R\&D Program of China (NO.2022ZD0160104).

{
    \small
    \bibliographystyle{ieeenat_fullname}
    \bibliography{main}
}
\clearpage

\section{Selection Process for Pretraining Subset: Focused on Cooking Videos}
For our pretraining process, we carefully selected a subset of HowTo100M~\cite{miech19howto100m} videos, leveraging available metadata to curate a tailored dataset. Focusing on cooking-related content, we filtered videos based on the "Recipes" category\_2 tag, ensuring relevance to recipe-based tasks. Within this subset, we further refined our selection by prioritizing videos within the top 30 rankings. This selection strategy aimed to capture cooking-related content, given its sequential nature and specificity in depicting step-by-step instructions. Cooking videos often present a structured flow, mirroring the requirements for dense video captioning tasks. Their explicit actions and clear events make them ideal for training models to capture and describe such events accurately. Additionally, these instructional videos offer rich visual and textual information, providing diverse examples pivotal for robust model training and the generation of precise captions for subsequent tasks.

\section{Prompt Engineering}
To streamline the process of deriving step-by-step event sequences from video subtitles, we utilize LLMs like LLAMA2~\cite{touvron2023llama2} and ChatGPT to generate prompts. Initially, our requests for prompts, such as '\textit{Task: write a prompt to send to LLM to extract the ordered steps of a recipe video subtitle. Example steps should resemble "coat the chicken pieces in the flour"},' resulted in step-by-step event captions that did not align with the concise, action-oriented format found in YouCook2~\cite{ZhXuCo18youcook2} captions.

To enhance this extraction process further, we manually refine the prompts to mirror the succinct nature of YouCook2 captions. Specifically, we emphasize extra regularization highlighting single-sentence clarity, exclusion of events unrelated to video actions, and emphasis on the sequential process. Additionally, after the instruction prompt, we append the original subtitle text followed by "\textbackslash nSteps:" to extract the step-by-step events without extra irrelevant words. The complete input for LLM should resemble "\textit{Prompt $[original\;subtitle]$ \textbackslash nSteps:}". This fine-tuning aims to closely align the extracted events with the concise and action-focused nature of YouCook2 captions, potentially shedding light on why pretraining shows better performance in YouCook2~\cite{ZhXuCo18youcook2} compared to ActivityNet~\cite{krishna2017anet}. Some potential prompt candidates are detailed in Table~\ref{tab:prompt}.

\begin{table*}[htbp]
    \centering
    \begin{tabular}{c|p{0.9\textwidth}}
        \toprule
        \textbf{Index} & \textbf{Prompt Description} \\
        \midrule
        1 & Task: Extract ordered steps from video subtitles, focusing on sequential process and action-oriented instructions. \\
        \midrule
        2 & Task: Please extract concise and action-oriented steps from video subtitles, ensuring a clear sequential arrangement of events. \\
        \midrule
        3 & Task: Generate steps from subtitles in a single-sentence, emphasizing clear actions, excluding non-action-related events. \\
        \midrule
        4 & Task: Extract concise and action-oriented steps or instructions from the video subtitles, focusing solely on the sequential process. Each step should be presented as a single sentence with clear actions. \\
        \midrule
        5 & Task: Extract concise and action-oriented steps or instructions from the video subtitles, focusing solely on the sequential process. Each step should be presented as a single sentence with clear actions. Exclude any steps that are not directly related to the actions in the video. \\
        \bottomrule
    \end{tabular}
    \caption{Examples of enhanced prompts for extracting step-by-step events from video subtitles.}
    \label{tab:prompt}
\end{table*}

\section{Additional ablation studies}
\paragraph{Effects of Pretraining Data Variation}
Our comparative experiments in Table \ref{tab:datasize} exploring varying pretraining data proportions on YouCook2 and ActivityNet reveal a notable trend. Increasing the pretraining data volume significantly boosts model performance on both event localization and caption generation. Strikingly, we witness a substantial performance jump on YouCook2 compared to ActivityNet. This suggests the instructional cooking focus within YouCook2 derives greater benefit from more pretraining data. The cooking domain aligns closely with the pretraining data, enabling more efficient learning. In contrast, ActivityNet's broader activity spectrum may need more diverse pretraining data to show similar improvements. This highlights the pivotal role of dataset characteristics and domain relevance in pretraining efficacy for dense video captioning. 

\begin{table}[tb]
    \centering
    \resizebox{0.48\textwidth}{!}{
    \begin{tabular}{c|c|cccccc}
    \toprule
        Dataset & Data  &  M & C & S & Rec. & Pre. & F1\\
        \midrule
        \multirow{5}{*}{YouCook2} & 0\% & 7.87 & 46.02 & 6.87 & 29.66 & 42.04 & 34.78 \\                    
                                  & 25\% & 8.64 & 52.47 & 7.89 & \textbf{31.87} & 43.25 & 36.70 \\
                                  & 50\% & 9.14 & 55.42 & 7.37 & 29.02 & 44.87 & 35.25 \\
                                  & 75\%  & 9.39 & \textbf{60.95} & \textbf{8.17} & 31.27 & 44.61 & \textbf{36.77} \\
                                  & 100\% & \textbf{9.41} & 59.35 & 7.97 & 30.80 & \textbf{45.13} & 36.61 \\
        \midrule
        \multirow{5}{*}{ActivityNet} & 0\% & 8.31 & 30.11 & 5.63 & 50.82 & 55.73 & 53.16 \\         
                                  & 25\% & 8.85 & 30.11 & 5.63 & 51.18 & 56.29 & 55.60 \\
                                  & 50\% & 8.72 & 30.00 & 5.70 & 53.29 & 57.93 & 55.51 \\
                                  & 75\% & 8.84 & \textbf{32.75} & \textbf{5.90} & \textbf{53.86} & 56.67 & 55.23 \\
                                  & 100\% & \textbf{8.93} & 31.89 & 5.85 & 53.14 & \textbf{58.31} & \textbf{55.61} \\
    \bottomrule
    \end{tabular}}
    \caption{Impact of varying pretraining data on dense video captioning performance on YouCook2 and ActivityNet.}
    \label{tab:datasize}
\end{table}

\paragraph{Pseudo Boundary Generation}
In our comparative study across different pseudo boundary generation schemes on the YouCook2 dataset, we evaluate the performance of models using uniform boundaries, Drop-DTW boundaries, and our proposed generation scheme in Table~\ref{tab:boundary_generation}. The uniform boundary approach, dividing the entire video into segments based on the number of captions, results in moderate performance. Surprisingly, the Drop-DTW method exhibits higher caption metric performance than uniform boundaries but notably underperforms in localization metrics. This discrepancy suggests that the noise present in the similarity matrix might hinder the accurate generation of boundaries using Drop-DTW. Conversely, our proposed scheme showcases superior performance in both localization and caption metrics, indicating the efficacy of our method in mitigating the impact of noise and improving overall performance in dense video captioning tasks.

\begin{table}[tb]
    \centering
    \resizebox{0.48\textwidth}{!}{
    \begin{tabular}{c|cccccc}
    \toprule
        Boundary  &  M & C & S & Rec. & Pre. & F1\\
        \midrule
        Uniform & 2.95 & 15.16 & 4.29 & 19.4 & 19.88 & 19.64 \\                    
                                   Drop-DTW & 3.21 & 17.37 & 3.86 & 14.09 & 17.26 & 15.51 \\
                                   Ours w/o STC  & 5.43 & 25.57 & 4.47 & 18.05 & 32.83 & 23.29 \\
                                   Ours w/ STC  & \textbf{5.88} & \textbf{28.04} & \textbf{4.47} & \textbf{19.72} & \textbf{35.43} & \textbf{25.34} \\
    \bottomrule
    \end{tabular}}
    \caption{Comparative analysis of pseudo boundary generation schemes for dense video captioning on YouCook2.}
    \label{tab:boundary_generation}
\end{table}

\paragraph{Few-shot Model Performance}
In our investigation of few-shot dense video captioning, we analyze model performance with varying amounts of fine-tuning data on the YouCook2 dataset, as detailed in Table~\ref{tab:few-shot}. Notably, as expected, models exhibit improved performance as more fine-tuning data is utilized, showcasing a positive correlation between data volume and model performance. Moreover, our comparison between models with and without pretraining reveals a substantial performance boost in pretraining-enabled models across all fine-tuning data proportions. Intriguingly, models with pretraining showcase significantly superior performance even with just half the fine-tuning data, surpassing the performance of models without pretraining. These results underscore the remarkable effectiveness of pretraining and our pseudo annotation generation method in enhancing model capabilities for few-shot dense video captioning tasks.
\begin{table}[t]
    \centering
    \resizebox{0.48\textwidth}{!}{
    \begin{tabular}{cc|cccccc}
    \toprule
         Pretrain & FT Data  &  M & C & S & Rec. & Pre. & F1\\
        \midrule
        \xmark & 25\% & 5.06 & 22.39 & 4.45 & 23.94 & 36.41 & 28.89 \\
        \cmark & 25\% & 7.81 & 46.69 & 7.17 & 28.93 & 41.08 & 33.95 \\
        \xmark & 50\% & 6.45 & 34.30 & 5.44 & 25.18 & 39.75 & 30.83 \\
        \cmark & 50\% & 8.60 & 55.73 & 7.84 & 30.14 & 42.73 & 35.35 \\
        \xmark & 75\%  & 7.26 & 40.49 & 6.31 & 27.81 & 40.60 & 33.01 \\
        \cmark & 75\%  & 9.11 & 59.09 & 7.86 & 29.97 & 44.54 & 35.83 \\
        \xmark & 100\% & 7.87 & 46.02 & 6.87 & 29.66 & 42.04 & 34.78 \\       
        \cmark & 100\% & \textbf{9.41} & \textbf{59.35} & \textbf{7.97} & \textbf{30.80} & \textbf{45.13} & \textbf{36.61} \\

    \bottomrule
    \end{tabular}}
    \caption{Impact of fine-tuning data and pretraining on few-shot dense video captioning performance in YouCook2. ``FT data'' represents the percentage of data used for fine-tuning.}
    \label{tab:few-shot}
\end{table}

\paragraph{Effects of top-$\hat{k}$ in Pseudo Boundary Generation}
In our investigation of top-$\hat{k}$ values' impact on pseudo boundary generation and subsequent model performance on YouCook2 and ActivityNet, detailed in Table~\ref{tab:pseudo_boundary-k}, contrasting trends emerge between the two datasets. Surprisingly, YouCook2 displays improved performance as the top-$\hat{k}$ value decreases, while ActivityNet exhibits a contrasting trend, showing deteriorating results with smaller top-$\hat{k}$ values for partial metrics. Our pseudo boundary generation scheme operates such that a larger top-$\hat{k}$ value leads to an inclusion of more frames, potentially unrelated to the current event caption. We hypothesize that YouCook2, known for its diverse events, likely has a higher frequency of events, thereby resulting in shorter event durations compared to ActivityNet. This difference in event duration might explain the disparate impact of top-$\hat{k}$ values on model performance between the two datasets.
\begin{table}[t]
    \centering
    \resizebox{0.48\textwidth}{!}{
    \begin{tabular}{c|c|cccccc}
    \toprule
         Dataset & $\hat{k}$  &  M & C & S & Rec. & Pre. & F1\\
        \midrule
        \multirow{4}{*}{YouCook2}
        &  15 & \textbf{5.91} & \textbf{30.89} & \textbf{5.00} & \textbf{20.56} & \textbf{35.92} & \textbf{26.15} \\
        &  20 & 5.79 & 27.04 & 4.72 & 18.99 & 35.89 & 24.84 \\
        &  25 & 5.35 & 25.86 & 4.71 & 17.8 & 33.24 & 23.18 \\
        &  30 & 4.62 & 21.88 & 4.48 & 17.26 & 30.05 & 21.93 \\
        \midrule
        \multirow{4}{*}{ActivityNet}
        &  15 & 7.21 & \textbf{18.40} & \textbf{4.76} & \textbf{43.96} & 61.50 & 51.27 \\
        &  20 & 7.30 & 17.52 & 4.69 & 43.46 & 63.68 & 51.66 \\
        &  25 & 7.32 & 17.25 & 4.71 & 43.22 & 64.74 & \textbf{51.84} \\
        &  30 & \textbf{7.45} & 16.50 & 4.71 & 42.29 & \textbf{65.68} & 51.45 \\

    \bottomrule
    \end{tabular}}
    \caption{Impact of top-$\hat{k}$ values on model performance in YouCook2 and ActivityNet.}
    \label{tab:pseudo_boundary-k}
\end{table}

\paragraph{Effects of Refinement Hyperparameters.}
In Table~\ref{tab:proposal}, our exploration of the impact of the number of selected proposals for merging new boundaries (3, 4, and 5 proposals) reveals an intriguing trend across both YouCook2 and ActivityNet. As the number of proposals increases, we observe a trade-off in metrics. Fewer proposals tend to select boundaries closely aligned with the current caption, suggesting a focus on relevance, while a larger count encompasses a more diverse selection, albeit with some proposals less related to the ongoing caption. This dual impact leads to a compromise in model performance, highlighting the detrimental effects when the number of proposals deviates too far from an optimal range.

\begin{table}[t]
    \centering
    \resizebox{0.48\textwidth}{!}{
    \begin{tabular}{c|c|cccccc}
    \toprule
         Dataset & $K$  &  M & C & S & Rec. & Pre. & F1\\
        \midrule
        \multirow{3}{*}{YouCook2}
        &  3 & 5.90 & 29.62 & 4.96 & 20.44 & \textbf{36.38} & \textbf{26.17} \\
        &  4 & 5.89 & 29.28 & 4.85 &19.85 & 35.67 & 25.51 \\
        &  5 & \textbf{5.91} & \textbf{30.89} & \textbf{5.00} & \textbf{20.56} & 35.92 & 26.15 \\
        \midrule
        \multirow{3}{*}{ActivityNet}
        &  3 & 7.33 & 15.86 & 4.67 & 42.21 & \textbf{65.79} & 51.43 \\
        &  4 & \textbf{7.45} & \textbf{16.50} & \textbf{4.71} & 42.29 & 65.68 & 51.45 \\
        &  5 & 7.35 & 16.27 & 4.68 & \textbf{42.77} & 65.54 & \textbf{51.76} \\

    \bottomrule
    \end{tabular}}
    \caption{Impact of selected proposals on boundary merging in YouCook2 and ActivityNet.}
    \label{tab:proposal}
\end{table}

\begin{table}[t]
    \centering
    \resizebox{0.48\textwidth}{!}{
    \begin{tabular}{c|c|cccccc}
    \toprule
         Dataset & Stages  &  M & C & S & Rec. & Pre. & F1\\
        \midrule
        \multirow{3}{*}{YouCook2}
        &  1 & \textbf{5.95} & \textbf{32.08} & 4.78 & 19.39 & 35.79 & 25.15 \\
        &  2 & 5.90 & 29.62 & 4.96 & \textbf{20.44} & \textbf{36.38} & \textbf{26.17} \\
        &  3 & 5.74 & 28.95 & \textbf{4.98} & 20.29 & 34.72 & 25.61 \\
        \midrule
        \multirow{3}{*}{ActivityNet}
        &  1 & 7.34 & 15.93 & 4.65 & \textbf{43.10} & 65.5 & \textbf{51.99} \\
        &  2 & 7.33 & 15.86 & \textbf{4.67} & 42.21 & \textbf{65.79} & 51.43 \\
        &  3 & \textbf{7.37} & \textbf{16.04} & 4.65 & 42.40 & 65.39 & 51.44 \\

    \bottomrule
    \end{tabular}}
    \caption{Impact of refinement stages on model performance in YouCook2 and ActivityNet.}
    \label{tab:stage}
\end{table}

In Table~\ref{tab:stage}, we examine the influence of refinement stages (1, 2, and 3 stages) and observe distinct trends across both YouCook2 and ActivityNet. On YouCook2, we observe a trend of declining caption performance with increasing refinement stages, accompanied by a marginal decrease in localization metrics with larger stage counts. Similarly, ActivityNet exhibits varying metric shifts with increased stages, suggesting that an excessive number of refinement stages may not consistently enhance performance and could lead to fluctuations in metric outcomes. This nuanced relationship emphasizes that excessive refinement may not always yield improved results and could elongate the training process.

\begin{figure}[t]
    \centering
    \begin{subfigure}[b]{0.95\linewidth}
        \includegraphics[width=\linewidth]{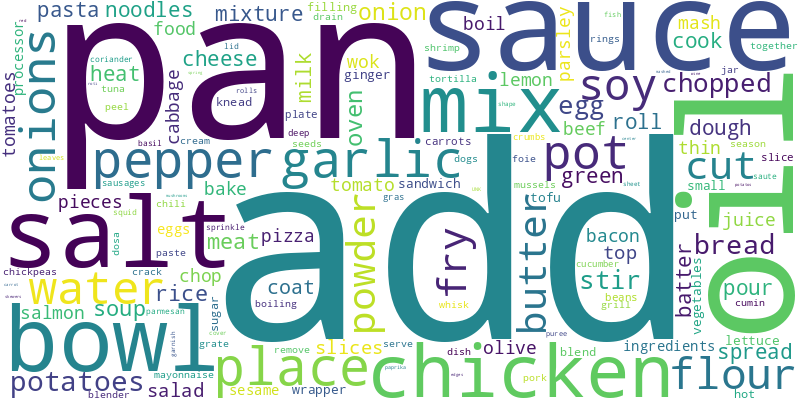}
        \caption{Without pretraining}
        \label{subfig:word_cloud_PDVC}
    \end{subfigure}
    \hfill
    \begin{subfigure}[b]{0.95\linewidth}
        \includegraphics[width=\linewidth]{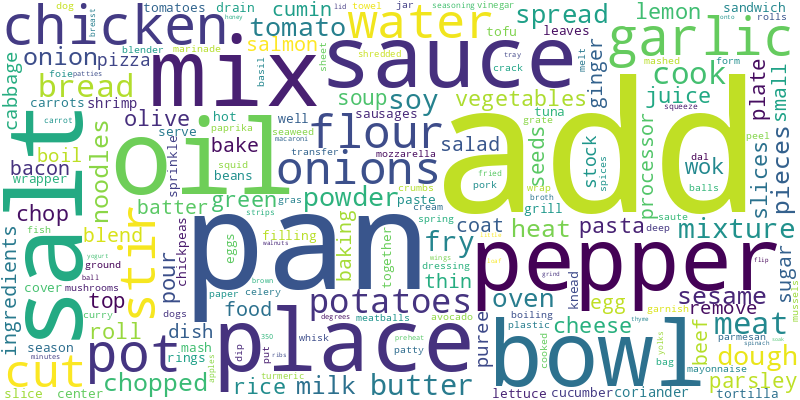}
        \caption{With pretraining}
        \label{subfig:word_cloud_DIBS}
    \end{subfigure}
    \caption{Word cloud of predicted captions from our DIBS w/ and w/o pretraining. A larger word indicates a higher frequency of its occurrence in the predicted captions.}
    \label{fig:word_clouds_comparison}
\end{figure}

\subsection{Visualization Results}

\paragraph{Does pretraining lead to diverse captioning?} In Figure~\ref{fig:word_clouds_comparison}, we compare word clouds of predicted captions from  DIBS model with and without our proposed pretraining. It can be seen that predicted captions after pretraining (Figure~\ref{subfig:word_cloud_DIBS}) show a richer vocabulary and broader concepts than those without pretraining (Figure~\ref{subfig:word_cloud_PDVC}), highlighting our method's success in diversifying caption generation.  The variation in word size and diversity after pretraining suggests significant improvements in the model's nuanced understanding, supporting our hypothesis on the benefits of pretraining for rich and diverse dense video captioning.



\begin{figure*}
    \centering
    \includegraphics[width=0.95\linewidth]{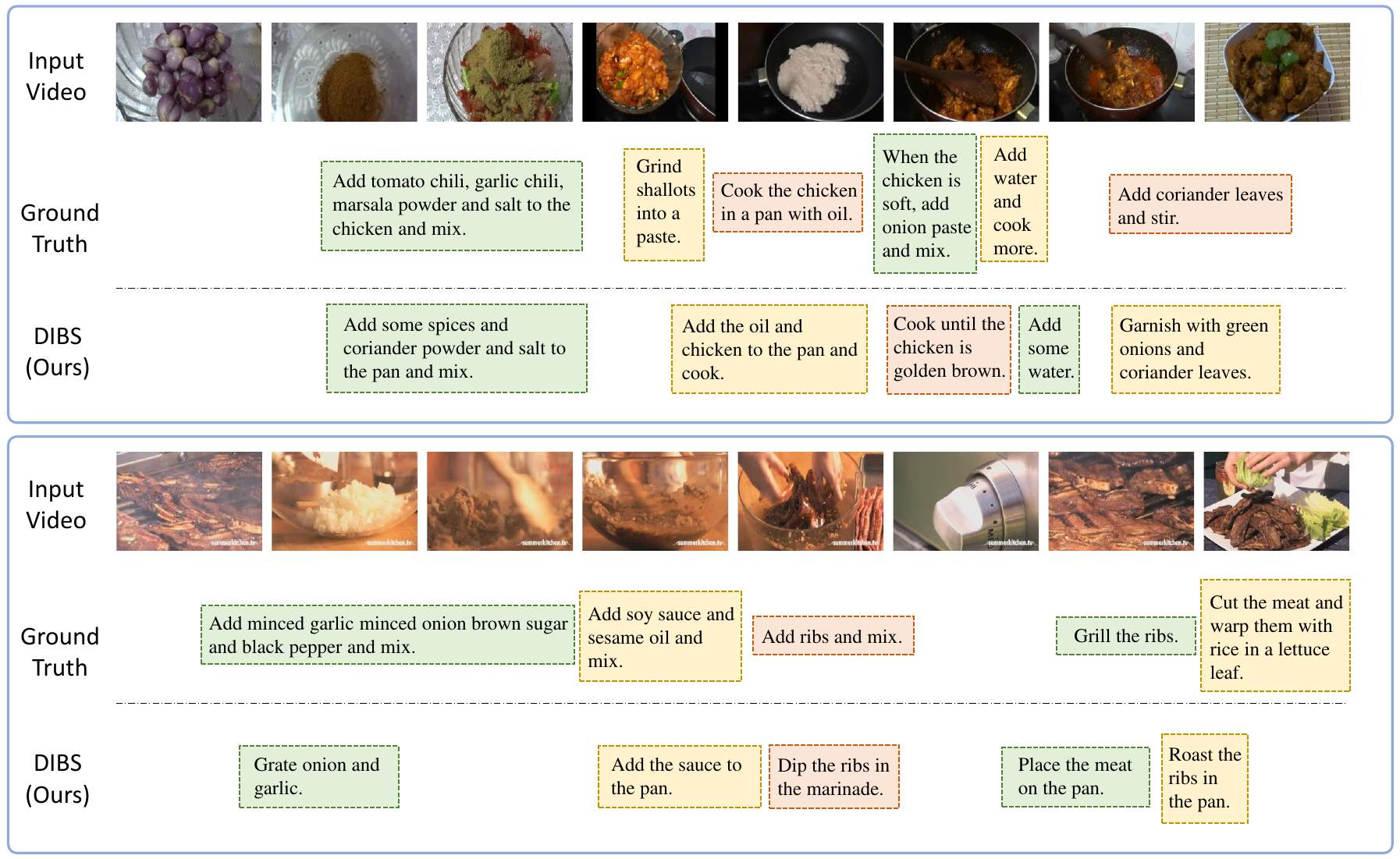}
    \caption{Visualization of qualitative results of our DIBS model. Video examples are selected from the validation set of YouCook2.}
    \label{fig:vis_yc2}
\end{figure*}

\begin{figure*}
    \centering
    \includegraphics[width=0.95\linewidth]{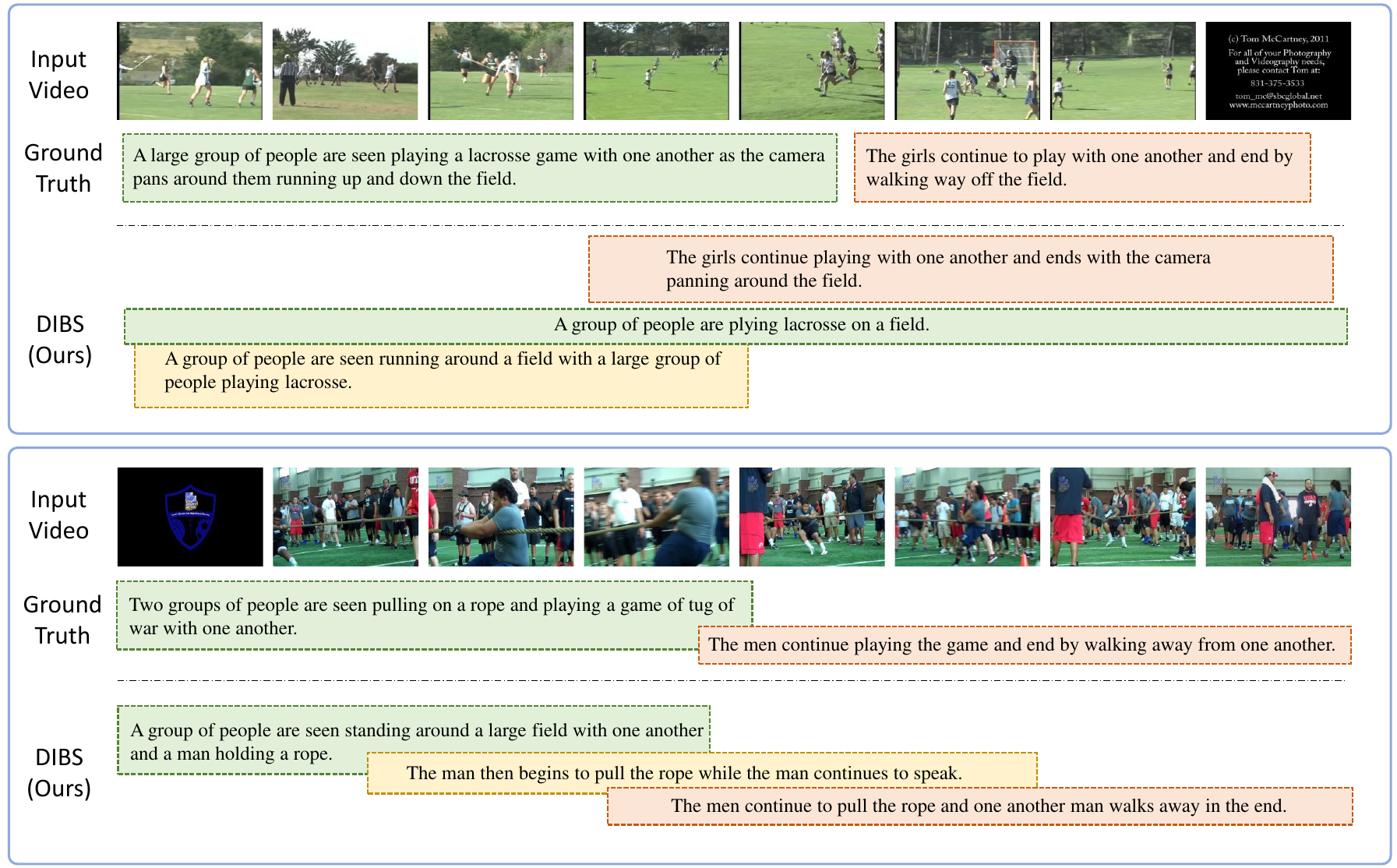}
    \caption{Visualization of qualitative results of our DIBS model. Video examples are selected from the validation set of ActivityNet.}
    \label{fig:vis_anet}
\end{figure*}

\paragraph{Qualitative Results} Finally, to further prove the effectiveness of our method, we present some qualitative results of DIBS on YouCook2 and ActivityNet datasets in Figure~\ref{fig:vis_yc2} and Figure~\ref{fig:vis_anet}, respectively. We can see from these figures that our DIBS could predict accurate event boundaries along with rich captions.

\end{document}